\documentclass[conference]{IEEEtran}
\IEEEoverridecommandlockouts
\usepackage{cite}
\usepackage{amsmath,amssymb,amsfonts}
\usepackage{algorithmic}
\usepackage{graphicx}
\usepackage{textcomp}
\usepackage[table]{xcolor}
\usepackage{url}
\usepackage{booktabs}
\def\BibTeX{{\rm B\kern-.05em{\sc i\kern-.025em b}\kern-.08em
    T\kern-.1667em\lower.7ex\hbox{E}\kern-.125emX}}
\begin{document}

\title{A 360-Degree Vision Dataset for Learning Yaw Control on GPS-Denied Micro-UAVs in Disaster-Response-Relevant Environments\\
\thanks{This work is funded by the German Ministry of Education and Research (BMBF), grant No. 13N16478, project ``E-DRZ: Establishment of the German Rescue Robotics Center'' (DRZ) \protect\url{https://www.rettungsrobotik.de}}
}

\author{\IEEEauthorblockN{Niklas Voigt}
\IEEEauthorblockA{\textit{Robotics Laboratory} \\
\textit{Westphalian University of Applied Sciences (WHS)}\\
Gelsenkirchen, Germany \\
niklas.voigt@w-hs.de}
\and
\IEEEauthorblockN{Hartmut Surmann}
\IEEEauthorblockA{\textit{Robotics Laboratory} \\
\textit{Westphalian University of Applied Sciences (WHS)}\\
Gelsenkirchen, Germany \\
hartmut.surmann@w-hs.de}
}

\maketitle

\begin{abstract}
This paper presents a novel data-driven approach to camera-based autonomy for micro-drones in GPS-denied, radio-challenging indoor environments. The target application is disaster and emergency response, where micro-UAVs can provide rapid situational awareness in hazardous settings such as firefighting and chemical, biological, radiological, and nuclear (CBRN) incidents while reducing risk for human responders. When the communication link is lost, the micro-drone uses a learned yaw controller to autonomously navigate toward open space, preserving onboard sensor data that would otherwise be lost with the vehicle. A custom micro-drone equipped with a 360-degree camera was used to record diverse industrial, underground, and training scenarios representative of communication-denied field operations. We introduce a preprocessing pipeline that converts equirectangular 360-degree footage into planar front views and dynamically generates image-label pairs for AI training. We then train and compare multiple convolutional neural network variants that predict a continuous yaw command from a single monocular view. Evaluation on a held-out test set confirms the feasibility of the learned yaw-prediction approach. A semi-autonomous real-world test further demonstrates the practicality of the method while revealing key failure modes, particularly reflections and glare.
\end{abstract}


\section{Introduction}

Unmanned aerial vehicles (UAVs) are increasingly used to support first responders and inspection teams because they can provide situational awareness without exposing human operators to direct risk~\cite{murphy2014disaster}.
More broadly, robots have been increasingly deployed in disaster and emergency response to provide situational awareness in hazardous environments while reducing risk to human responders.
Seminal work in disaster robotics has highlighted the importance of robust perception and autonomy under conditions such as GPS denial, communication loss, and environmental clutter, where human teleoperation may be unreliable or impossible~\cite{murphyrobotfailures}.

\begin{figure}[!t]
\centering
\includegraphics[width=0.32\columnwidth]{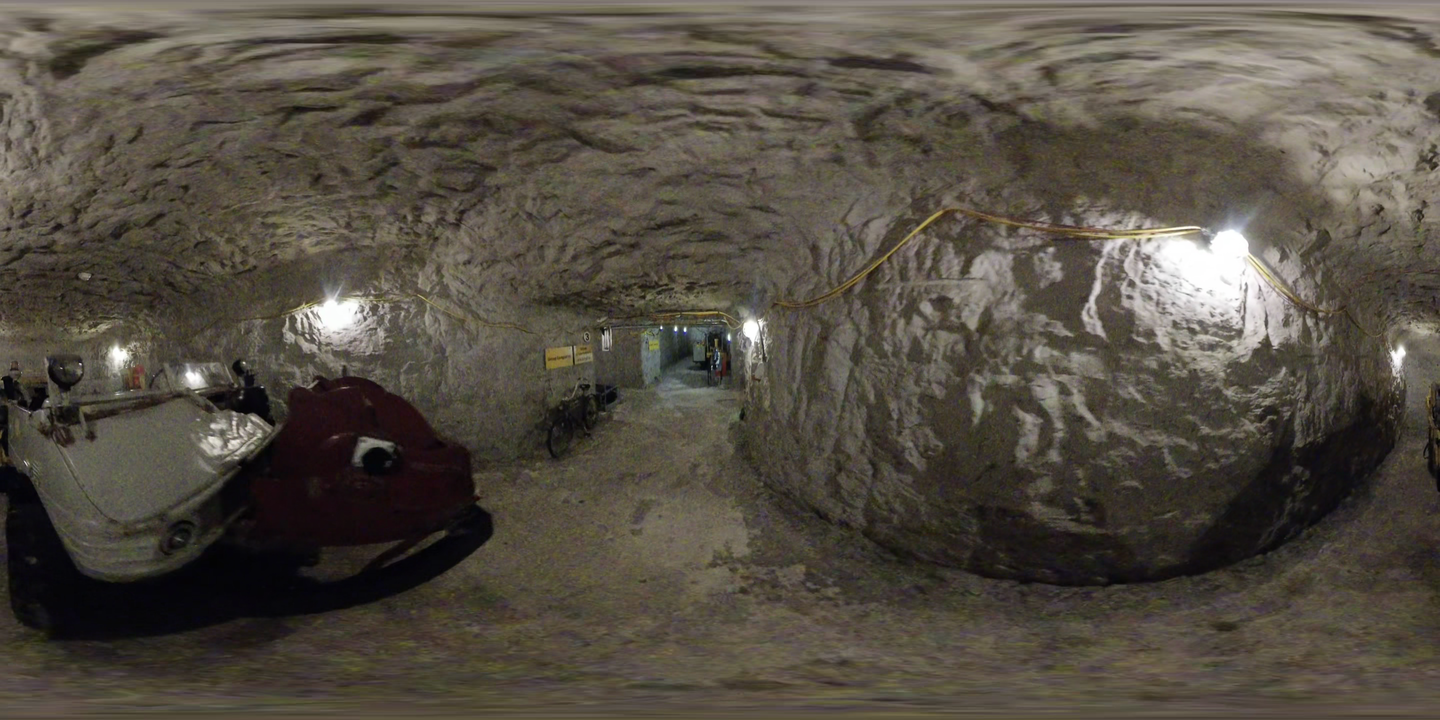}\hfill
\includegraphics[width=0.32\columnwidth]{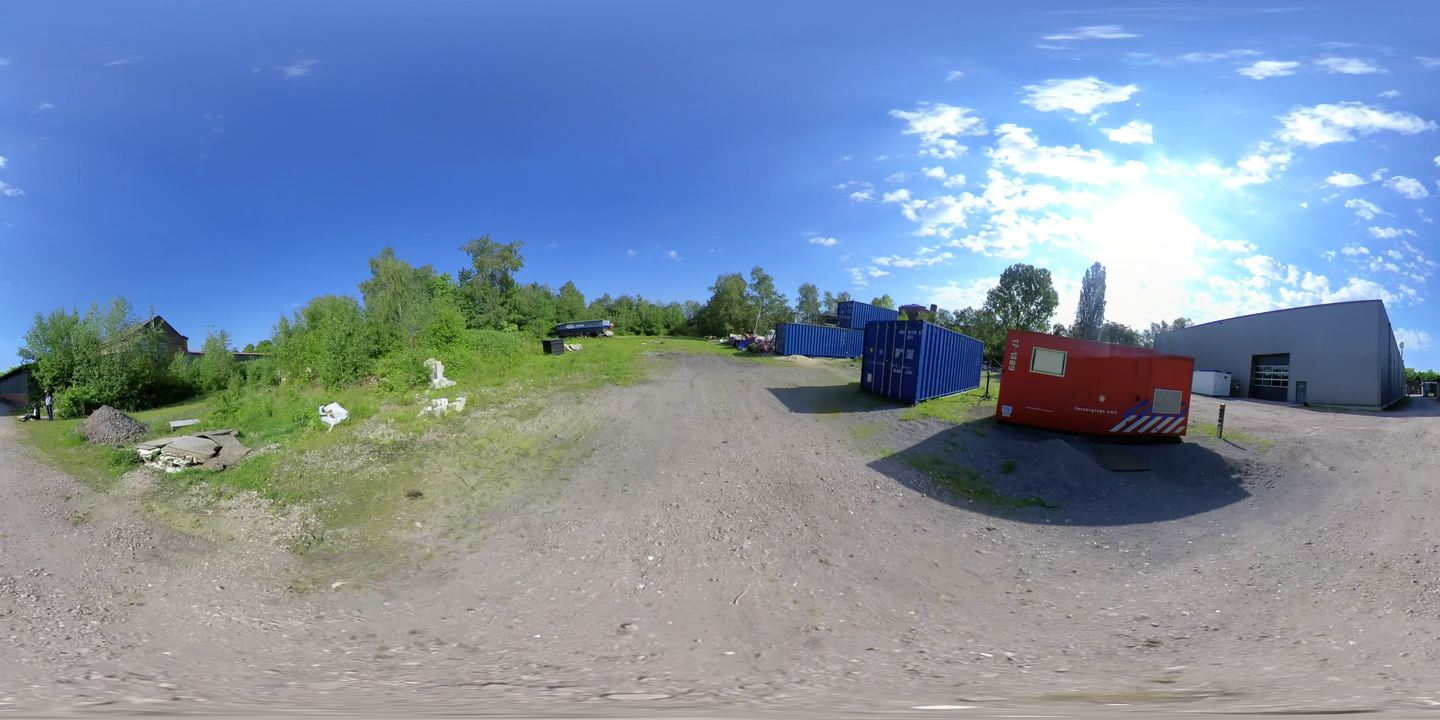}\hfill
\includegraphics[width=0.32\columnwidth]{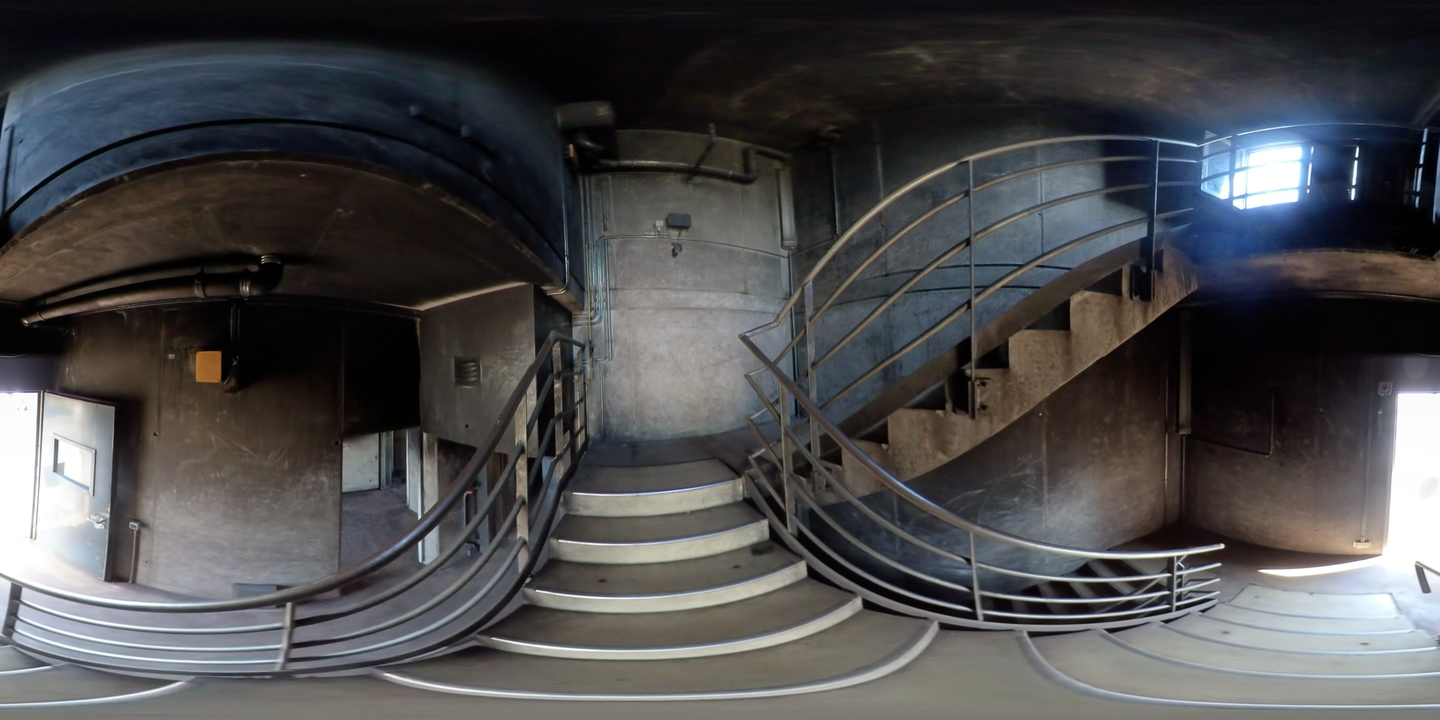}

\vspace{0.35em}
\includegraphics[width=0.32\columnwidth]{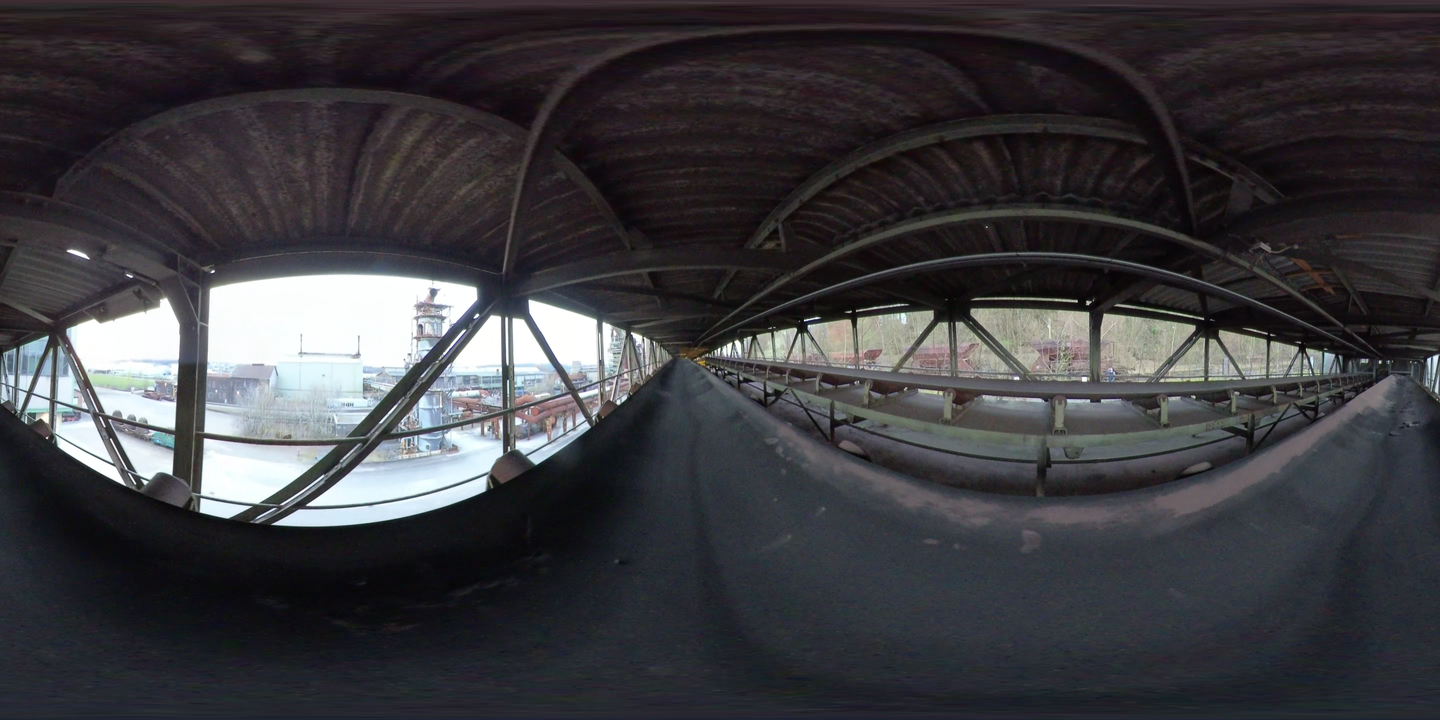}\hfill
\includegraphics[width=0.32\columnwidth]{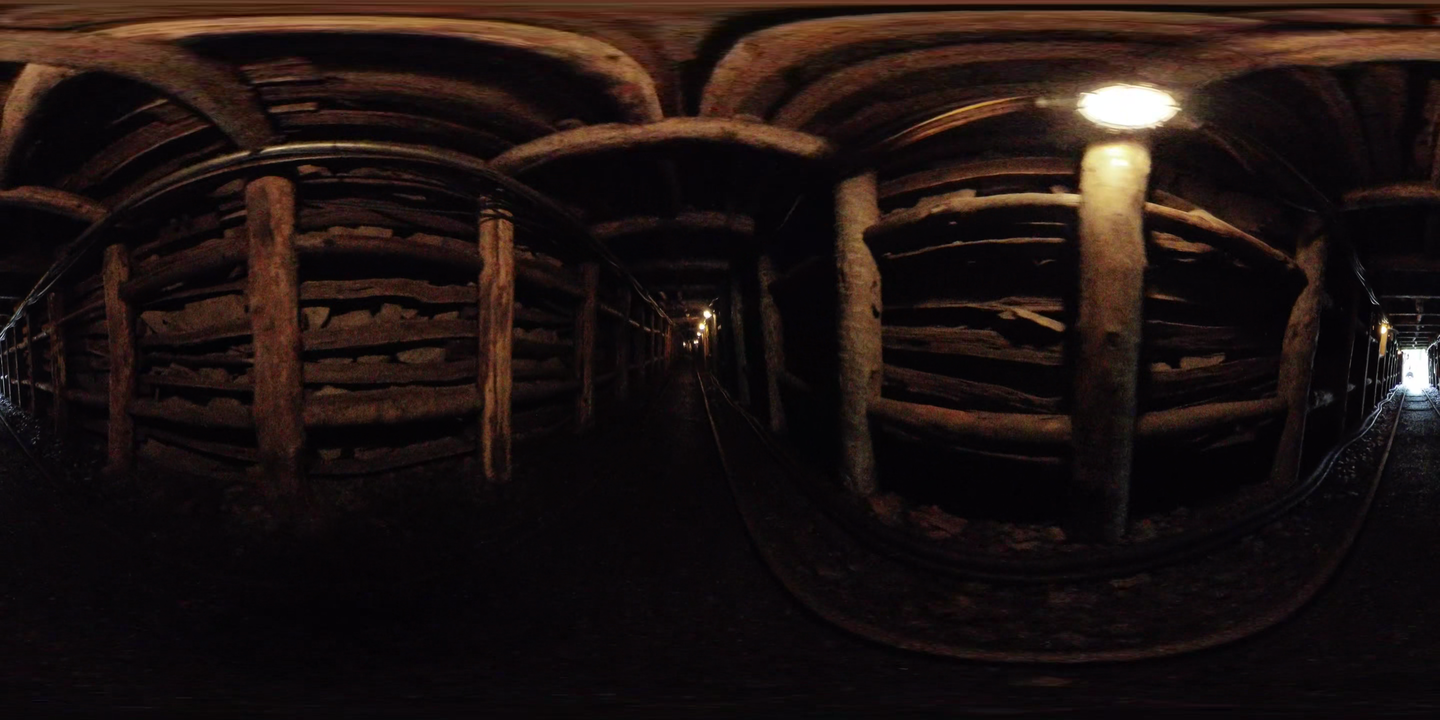}
\includegraphics[width=0.32\columnwidth]{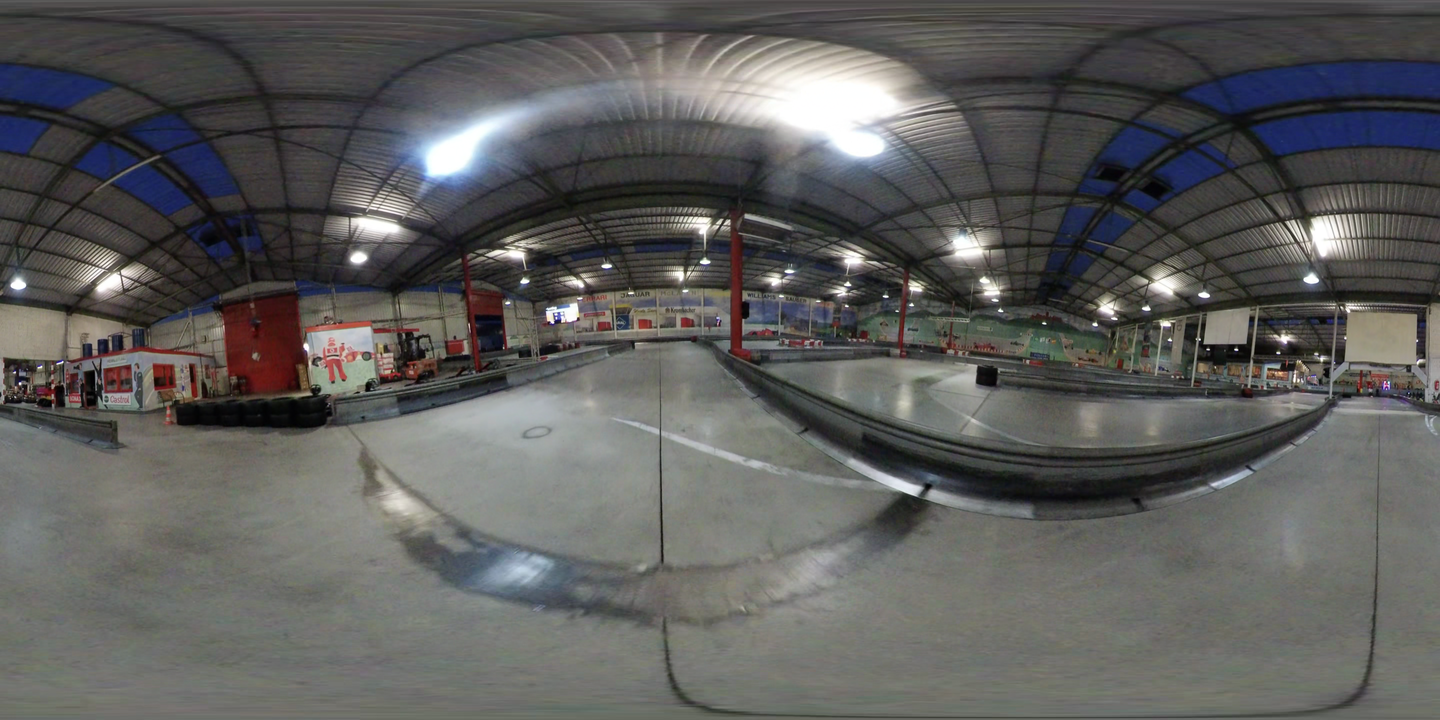}

\vspace{0.35em}
\includegraphics[width=0.32\columnwidth]{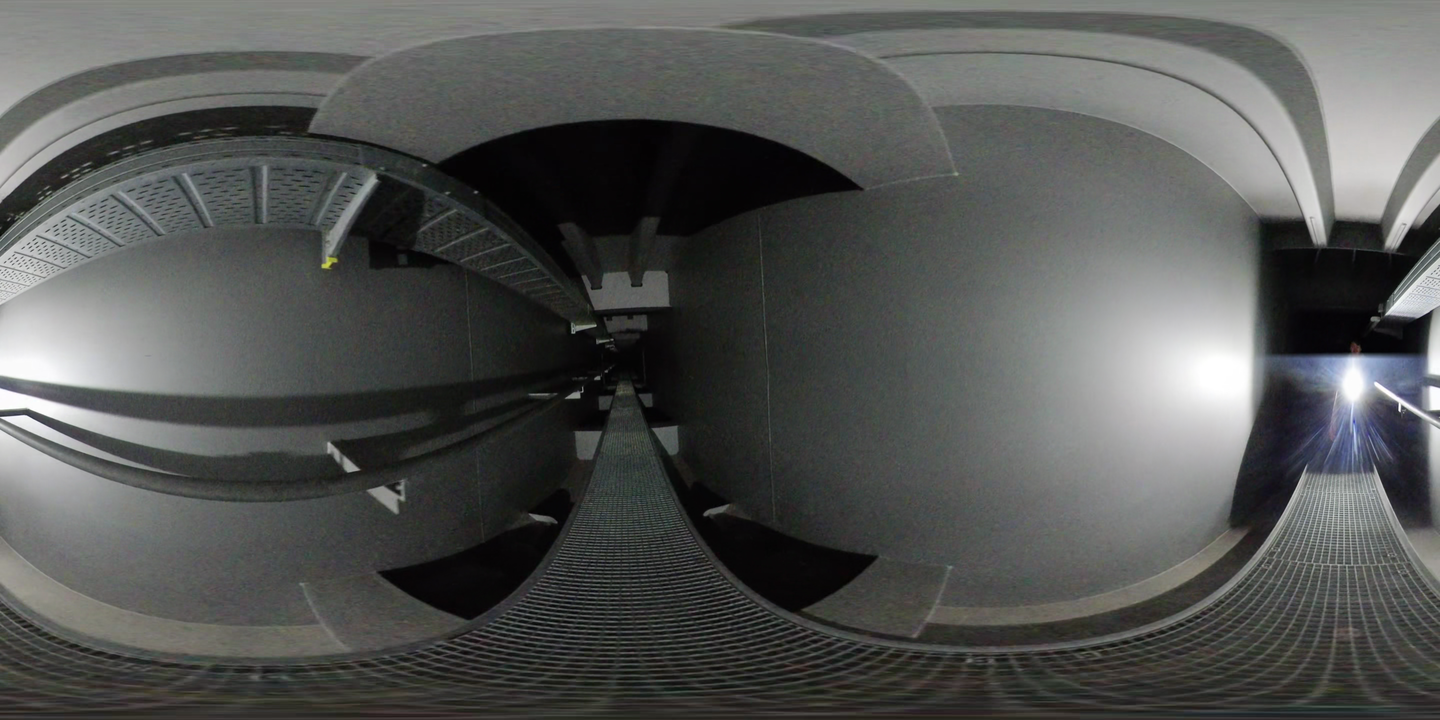}\hfill
\includegraphics[width=0.32\columnwidth]{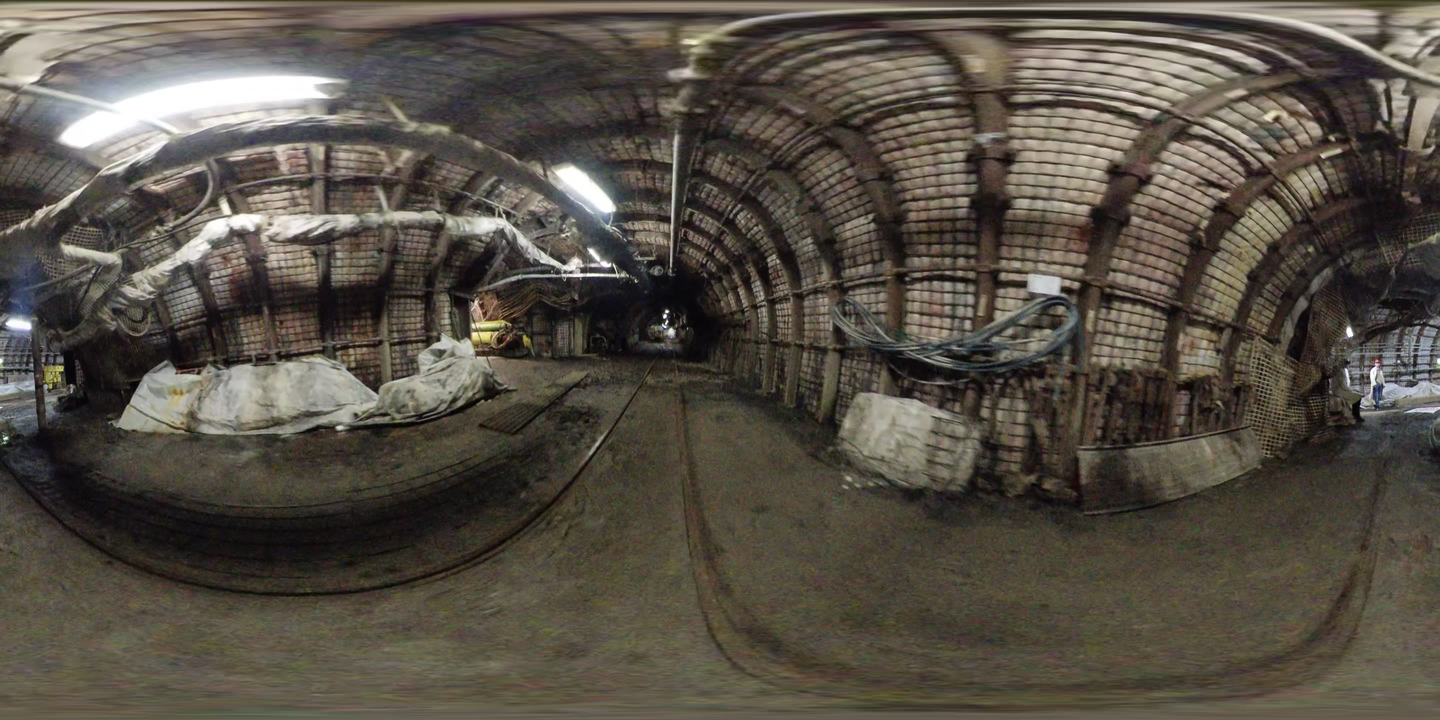}\hfill
\includegraphics[width=0.32\columnwidth]{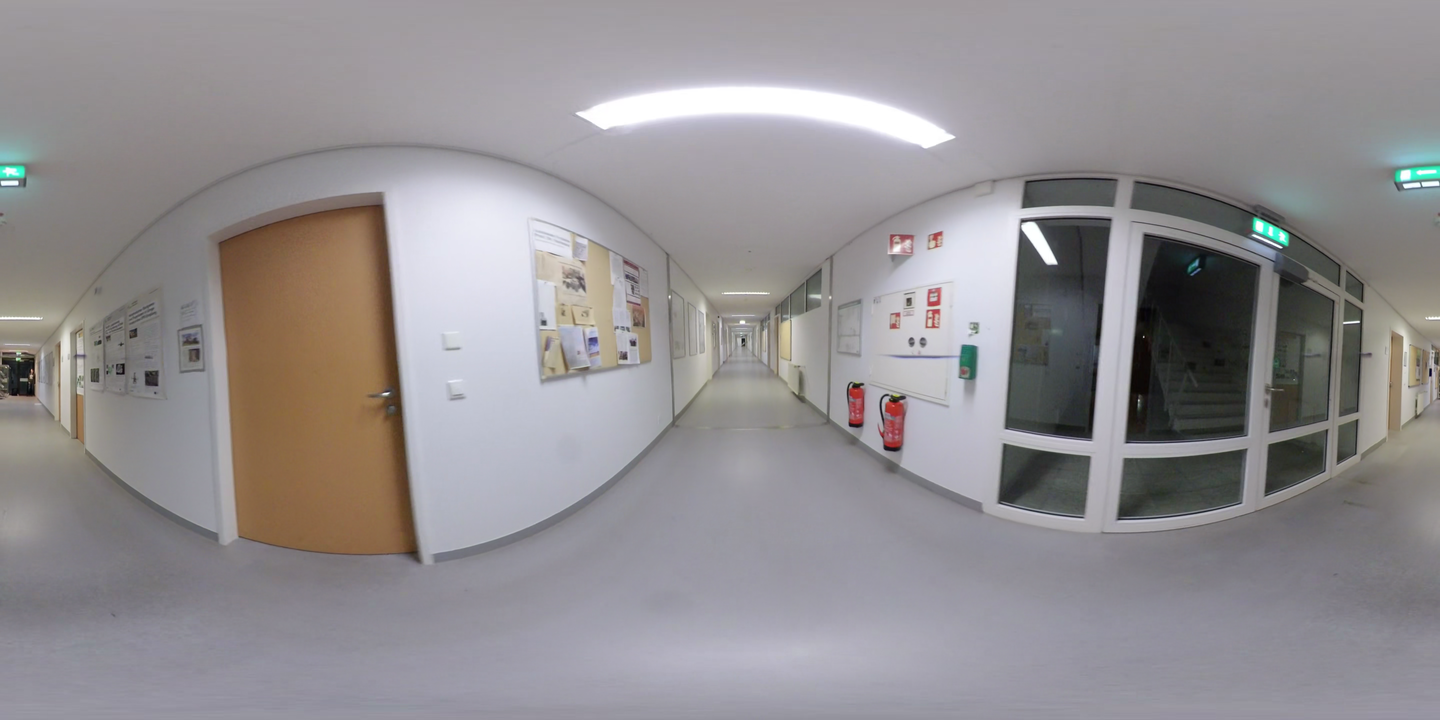}
\caption{Dataset overview: A 3×3 grid of equirectangular 360° thumbnails showing diverse GPS-denied disaster-response-relevant environments, including mines, fire training facilities, industrial sites, bridge passages, and outdoor areas.}
\label{fig:dataset-thumbnails}
\end{figure}

Micro-UAVs are particularly suited for reconnaissance in emergency response because they can be deployed quickly, reduce risk to humans, and can be treated as expendable assets if needed~\cite{einsaetzedrz}.
However, real deployments frequently occur in GPS-denied, cluttered, and communication-challenging environments such as industrial facilities, tunnels, and basements, where radio links degrade due to shadowing and attenuation~\cite{drz1}.
If the control link drops, a micro-drone that can autonomously stabilize and search for a safe exit path can substantially increase the operational value of such systems~\cite{DBLPTowardsResilient}.
This recovery capability addresses a concrete operational need by maintaining controlled flight and retreating from communication-denied regions, thereby preserving reconnaissance data (imagery, thermal maps, or gas sensor readings) that would otherwise be lost with the vehicle.
Such data recovery is often the primary mission objective in disaster response, where situational awareness can directly inform tactical decisions for human responders.

In addition to learning-based navigation, recent work has investigated the use of UAVs equipped with 360$^\circ$ cameras and neural scene learning techniques to improve situational awareness and 3D reconstruction in disaster environments~\cite{roblab}.

Many autonomous navigation stacks rely on active range sensing (e.g., LiDAR) for mapping and localization~\cite{ren2025safety}.
For micro-drones, however, such sensors can be too heavy, power hungry, or costly, motivating camera-only approaches that better align with payload, power, and cost constraints~\cite{liu2024omninxt}.
Vision-based learning approaches can exploit appearance cues such as corridor structure, openings, and obstacles to predict navigation commands when explicit mapping sensors are infeasible.

In this work, the learned yaw controller is designed as a pragmatic recovery mechanism: when the communication link is lost, it enables the micro-drone to navigate toward open space while preserving onboard sensor data that would otherwise be lost with the vehicle.
This paper therefore focuses on learning a direct mapping from a monocular front view to a continuous relative yaw control command.

The main contributions are:
\begin{itemize}
  \item a real-world dataset recorded with a custom micro-drone
equipped with a 360$^\circ$
camera covering diverse
indoor and outdoor scenarios, together with a preprocessing pipeline that converts equirectangular footage into
planar views and generates image–label pairs on-the-fly
to reduce memory footprint;
  \item an empirical comparison of multiple CNN variants for
yaw regression;
  \item a semi-autonomous real-world validation that reveals
practical strengths and limitations.
\end{itemize}

\section{Related Work}                      

Early learning-based navigation pipelines often used classification to select coarse steering directions.
Giusti \emph{et al.}~\cite{cnnpaper} trained a CNN to classify trail direction into three classes (turn left / go straight / turn right) based on monocular images and demonstrated outdoor trail following in   
forest environments.                                                                                                                                                                                             
Pearson \emph{et al.}~\cite{pearsonOptimized} proposed architectural and data-handling refinements (e.g., activation choices and dataset restructuring) and reported improved classification accuracy.           

In contrast to coarse direction classification, Samy \emph{et al.}~\cite{samy2019dronepathfollowinggpsdeniedenvironments} formulated yaw prediction as a regression problem and used a CNN-based pipeline trained
on synthetic data to follow waypoint paths in GPS-denied environments.
Motivated by these works, we focus on real-world indoor and outdoor data and continuous yaw regression for micro-drone navigation.                                                                               

More generally, our approach can be categorized as end-to-end imitation learning (also referred to as behavior cloning), where a policy is trained via supervised learning to directly map visual observations to
control commands based on demonstrated behavior.                                                                                                                                                                
Such approaches have been widely used in mobile robotics and autonomous driving due to their simplicity and suitability for real-time deployment, particularly when explicit mapping or planning is              
infeasible~\cite{nvidiaDBLP}.                                                                                                                                                                                    

In contrast to reinforcement-learning-based navigation, imitation learning enables direct use of real-world flight data without hand-crafted reward functions or online exploration — properties that have       
motivated its adoption in safety-critical UAV tasks such as obstacle avoidance~\cite{loquercioDroNet}, low-altitude MAV trail navigation~\cite{smolyanskiyTrailNet}, and agile flight in                         
clutter~\cite{loquercioWild}, and that are particularly relevant in disaster-response settings where online trial-and-error is not admissible.

Recent deep-reinforcement-learning approaches such as DCE~\cite{kulkarniDCE} and VDS-Nav~\cite{dangVDSNav} predict continuous 3D velocity and yaw-rate commands from depth images for closed-loop navigation in
cluttered environments, bridging the sim-to-real gap via a real-data-augmented latent encoder and a depth-based barrier reward, respectively. Their yaw-rate choice fits their task: the policy itself closes the
dynamics loop at sensor rate against a low-level velocity controller.

Our setting differs: a directional-guidance recovery component on a visual-inertial-stabilized platform, where the onboard stabilizer handles attitude and velocity tracking. The learning task
therefore reduces to identifying \emph{where} the recovery direction lies in a single monocular RGB view; the angle-to-rate gain is a downstream controller decision. An absolute yaw angle is the natural target
because each label is obtained geometrically as the extraction angle of a perspective view from an equirectangular panorama — a static, single-frame quantity. Deriving a yaw rate would require either
differentiating consecutive poses (coupling the network to the recording platform's dynamics) or applying an arbitrary gain that the controller can apply equivalently downstream.                               

Direct quantitative comparison with prior work is challenging because existing indoor navigation datasets (e.g., outdoor trail-following~\cite{cnnpaper}) target different environments and sensor               
configurations, and no established benchmark exists for yaw regression in GPS-denied disaster-proxy settings.
Our classification sanity check (Section~VI-A) confirms that the dataset and generator produce learnable signals consistent with prior three-class approaches, while the regression formulation enables
finer-grained control.

\section{System Overview}

This section describes the hardware platform used for data collection and the operational concept that motivates the learned yaw controller.

\subsection{Platform and Sensor Setup}

Data were collected primarily using a custom-built micro-drone with dimensions $200\,\mathrm{mm} \times 270\,\mathrm{mm}$ and a mass of $425\,\mathrm{g}$.
A modified BetaFPV SMO 360 camera (a stripped-down variant of the Insta360 ONE~R 360 module optimized for FPV applications) was integrated into the frame.
Each lens records at $2880 \times 2880$ pixels; after stitching, videos have a resolution of $5760 \times 2880$ at $\approx 30\,\mathrm{fps}$.

A key property of the dataset is that the recorded 360$^\circ$ video is unobstructed: the drone itself is not visible in the stitched video frames.
In typical 360$^\circ$ recordings from drone-mounted cameras, up to 33\% of the panoramic sphere is occluded by the platform itself~\cite{roblab}.
The mechanical integration used here, based on the invisible-drone platform introduced in~\cite{roblab}, eliminates this self-occlusion and makes the full panoramic sphere available for planar view extraction and dynamic dataset generation.
This is a structural prerequisite for the back-to-front mapping described in Section~IV-E, which doubles the effective training data by reusing the rear hemisphere of each frame.
In addition, the raw recordings contain synchronized inertial measurements (IMU) stored in metadata, which can be used to correct noisy labels and to improve robustness analysis in future work.

\begin{figure}[t]
\centering
\includegraphics[width=0.9\columnwidth]{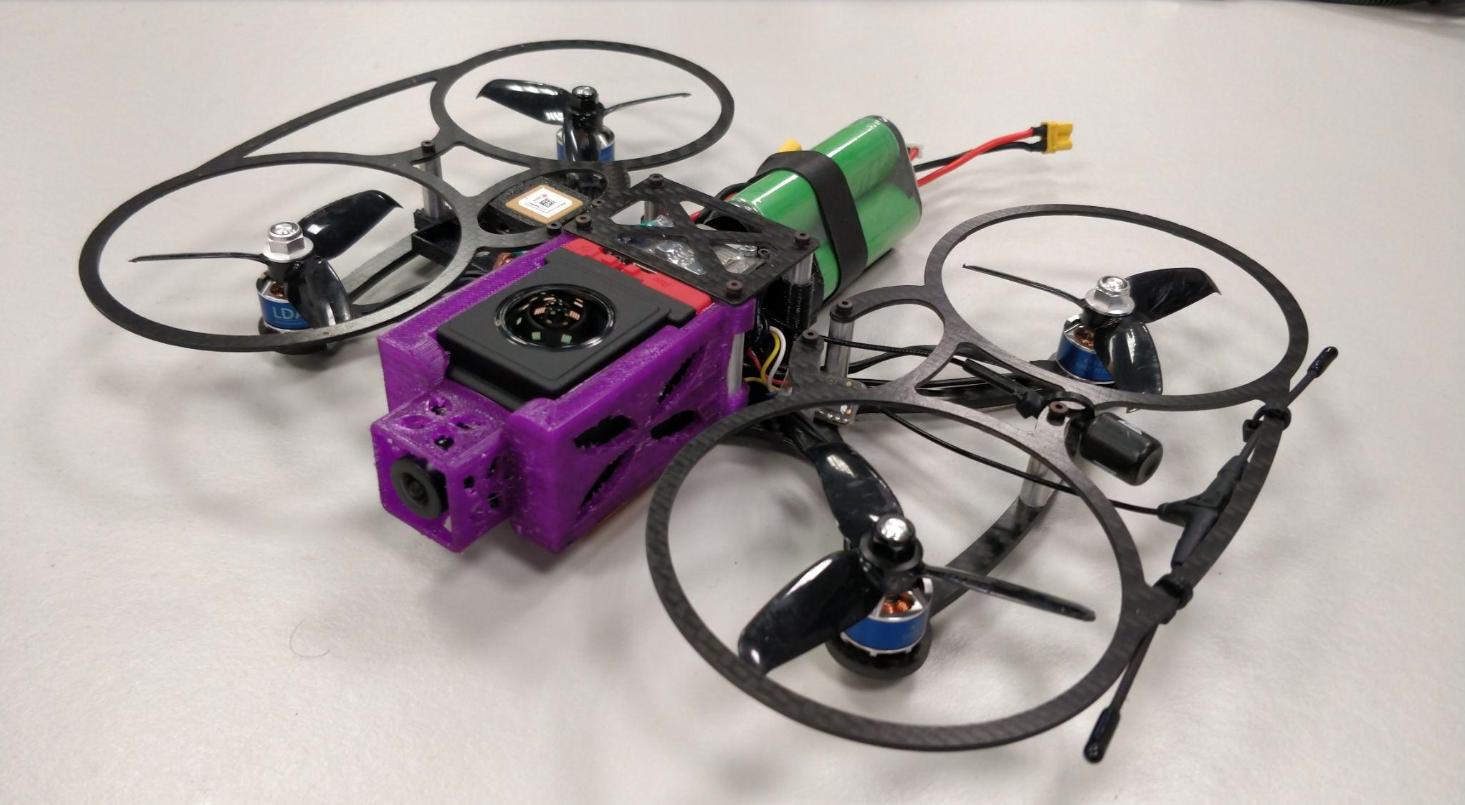}
\caption{Custom micro-drone platform with integrated 360$^\circ$ camera used for dataset recording.}
\label{fig:platform}
\end{figure}

\subsection{Operational Concept}

The learned yaw controller is designed as a recovery component: when the communication link to the operator is lost, the micro-drone uses its front-facing camera view to predict a yaw command that steers it toward open, traversable space.
By maintaining controlled flight and retreating from communication-denied regions, the platform preserves reconnaissance data that would otherwise be lost with the vehicle.
We emphasize that this yaw controller is an enabling component intended for integration with conservative safeguards (human-in-the-loop decision authority, geofencing, speed limits, and emergency stop behaviors) before operational deployment.

The learned yaw controller assumes that low-level state estimation and position hold are provided by a visual-interial odometry (VIO) pipeline such as OpenVINS or VINS-Fusion~\cite{openvins,vinsfusion}.
These systems fuse IMU and camera data to maintain stable hover and altitude hold even in GPS-denied environments.
Given this foundation, the drone executes a constant forward velocity while the CNN provides only the yaw correction command.
This separation of concerns restricts the learning problem to directional guidance---a deliberate design choice that exploits the maturity of VIO-based stabilization and avoids conflating flight control with navigation learning.

\section{Dataset and Preprocessing Pipeline}

This section details the recorded scenarios, the stitching workflow, and the dynamic dataset generator that converts 360$^\circ$ footage into planar image--label pairs for training.

\subsection{Recorded Scenarios}

Recordings were performed in multiple industrial and indoor locations, including underground structures and training facilities, with the goal of representing environments relevant to emergency response operations.
This aligns with prior work and practical deployments of aerial robots in firefighting and hazardous-material scenarios~\cite{drz9597869}.
While actual post-disaster environments (collapsed structures, debris fields) were not accessible for this study, the recorded industrial and underground scenarios share key perceptual challenges with disaster-affected structures: near-complete darkness (underground mine), heavy soot deposits and high dynamic range lighting (fire training facility), confined and degraded structural elements (industrial heritage sites), and narrow metallic passages (bridge inspection).
We acknowledge that true disaster scenes may present additional challenges such as dynamic smoke, irregular debris geometry, and moving obstacles.
In some locations, flight operations were not legally permitted; therefore, additional sequences were recorded manually using a tripod setup.

\subsection{Stitching and Alignment}

Raw dual-fisheye recordings were stitched into equirectangular videos and trimmed to remove takeoff/landing segments.
Because the camera is mounted with a fixed offset relative to the drone body, an additional yaw alignment is applied.

\begin{figure}[t]
\centering
\includegraphics[width=0.9\columnwidth]{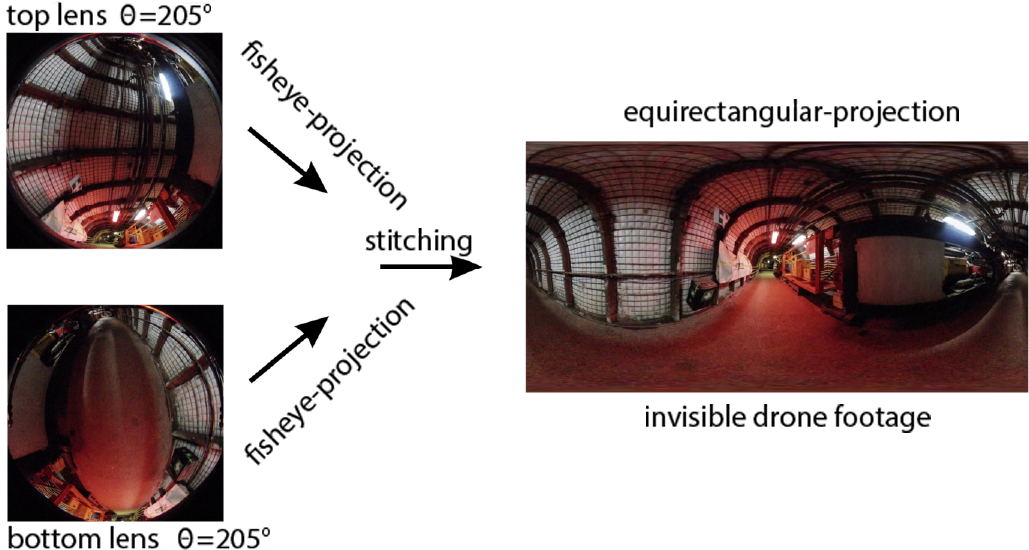}
\caption{Stitching pipeline from dual-fisheye camera views to an equirectangular representation.}
\label{fig:stitching}
\end{figure}

\subsection{Dataset Size and Splits}

From the raw videos, we manually selected representative sequences and extracted every 10th frame, corresponding to an effective sampling rate of approximately $3$\,fps.
This yielded the equirectangular frames summarized in Table~\ref{tab:dataset-summary}.
Since the primary deployment target is the autonomous micro-drone, the test set is composed exclusively of drone recordings, while the manual recordings serve as training data to broaden scenario diversity and act as a form of data augmentation.

\subsection{Dynamic Dataset Generator}

Instead of exporting a static set of planar images, we employ a dynamic dataset generator that produces planar input views and corresponding yaw labels at training time.
By densely sampling planar crops from equirectangular images, this approach acts as a strong data multiplier while reducing storage overhead and enabling controlled view sampling and consistent augmentation.

\begin{figure}[h]
\centerline{\includegraphics[width=0.9\columnwidth]{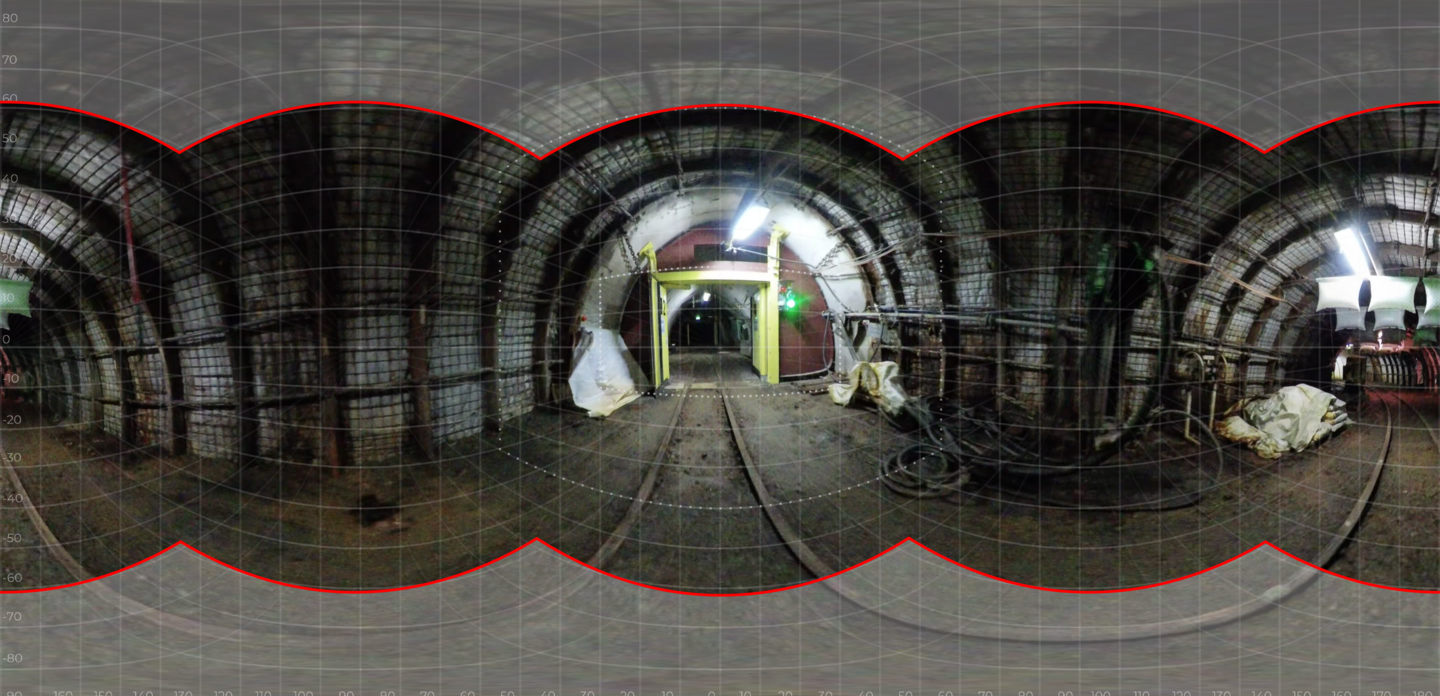}}
\caption{Equirectangular frame with the horizontal band used for planar view extraction (fixed pitch).}
\label{fig:equi-horizont}
\end{figure}

We start from stitched equirectangular frames and extract planar projections that emulate a conventional forward-facing camera.
To match the expected flight attitude during data collection, we fix the pitch to a level-flight configuration and restrict the 360$^\circ$ sphere to a horizontal band (Fig.~\ref{fig:equi-horizont}).

\begin{figure}[htbp]
\centerline{\includegraphics[width=0.9\columnwidth]{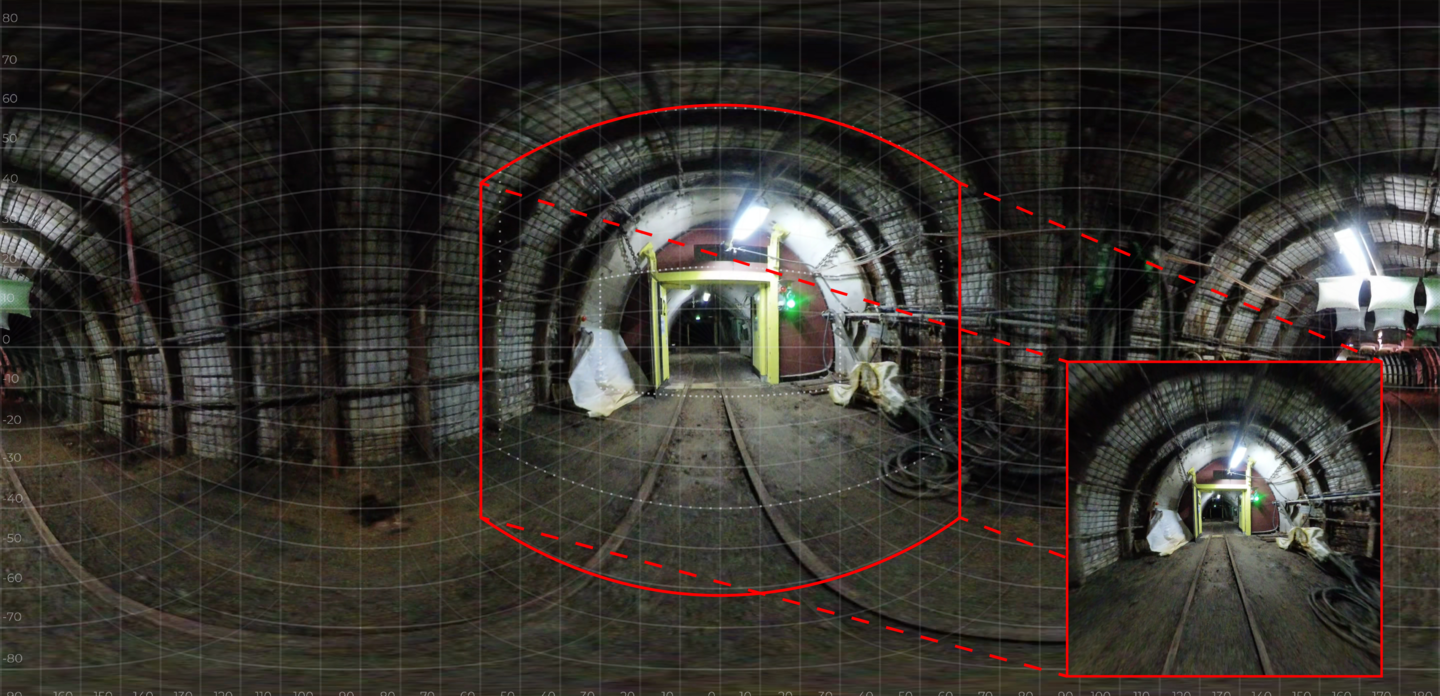}}
\caption{Example of extracting a planar view from an equirectangular frame.}
\label{fig:equi-pers}
\end{figure}

\begin{figure}[t]
\centerline{\includegraphics[width=0.8\columnwidth]{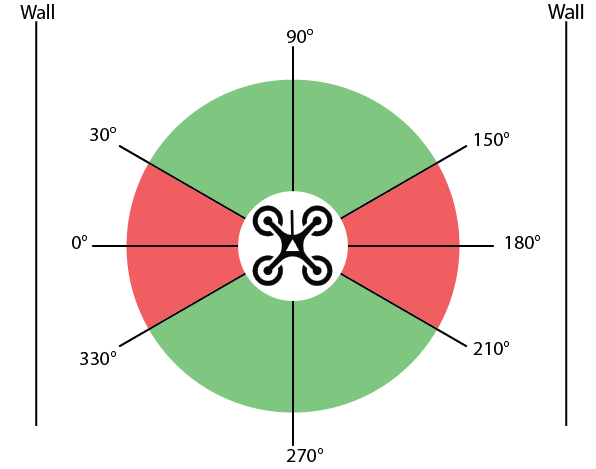}}
\caption{Visualization of yaw angles in the equirectangular image with respect to the drone (top view). Green regions mark the used field of view, while red regions indicate ambiguous boundary views that can confuse the network during training.}
\label{fig:winkelderdrohne}
\end{figure}

Planar images are extracted with a field of view (FOV) of $90^\circ$ and an initial resolution of $600\times 600$ px.
For training, we downsample to $101\times 101$ px to match the baseline network~\cite{cnnpaper} input size.
An example of the geometric transformation is shown in Fig.~\ref{fig:equi-pers}.
The yaw sampling regions are illustrated in Fig.~\ref{fig:winkelderdrohne}.

\subsection{On-the-Fly Pair Generation and Label Mapping}

The ground-truth yaw label for each planar view is derived directly from its 
extraction angle within the equirectangular frame.
By construction, the center of the equirectangular image at $\theta = 90^\circ$ 
corresponds to the drone's instantaneous heading 
(cf.\ Fig.~\ref{fig:winkelderdrohne}), as ensured by the yaw alignment described 
in Section~IV-C. During data collection, the drone was manually flown along 
recovery paths leading out of each scenario (e.g., toward the exit of a tunnel 
or corridor), with its yaw continuously aligned to the path direction~\cite{cnnpaper}. The 
centerpoint of every equirectangular frame therefore coincides with the 
demonstrated recovery heading at that point in time.
A planar view extracted at angle $\theta$ relative to this center thus directly 
encodes the yaw offset between an arbitrary viewing direction and the 
ground-truth recovery heading.

For each equirectangular source frame, the generator produces $N = 10$ planar 
views at yaw angles randomly sampled from the front range $[30^\circ, 150^\circ]$.
The extraction angle $\theta_{\mathrm{extract}}$ directly serves as the 
ground-truth label: a view extracted at $\theta_{\mathrm{extract}} = 60^\circ$ 
represents a perspective where the correct heading ($90^\circ$) lies $30^\circ$ 
to the right of the image center, requiring a rightward yaw correction. 
Conversely, $\theta_{\mathrm{extract}} = 120^\circ$ implies a $30^\circ$ 
leftward correction.

For drone recordings, we additionally sample $N$ views from the back 
range $[210^\circ, 330^\circ]$ and map them onto the front range via 
$\theta_{\mathrm{label}} = 360^\circ - \theta_{\mathrm{extract}}$. For example, 
a view extracted at $\theta_{\mathrm{extract}} = 210^\circ$ is relabeled as 
$\theta_{\mathrm{label}} = 150^\circ$, simulating the equivalent frontal 
perspective. This effectively doubles the number of training samples per frame 
while maintaining consistent label semantics.

Networks operate on a normalized output range. Let $t$ be the yaw label in 
degrees after applying the back-to-front mapping, where $t \in [30^\circ, 150^\circ]$.
We map it to the normalized range $y \in [-1, 1]$ via
\begin{equation}
  y = 2 \cdot \frac{t - 30^\circ}{120^\circ} - 1.
\end{equation}
For reporting and real-world control, the inverse mapping is applied to 
transform the network output back to degrees.

\begin{table}[htbp]
\caption{Dataset Summary (After Preprocessing)}
\label{tab:dataset-summary}
\centering
\begin{tabular}{lr}
\toprule
\textbf{Property} & \textbf{Value} \\
\midrule
Scenarios & 10 \\
Raw videos (drone / manual) & 156 (46 / 110) \\
Total duration & $2\,\mathrm{h}\,07\,\mathrm{min}\,49\,\mathrm{s}$ \\
\midrule
\multicolumn{2}{l}{\textbf{Equirectangular frames}} \\
\quad Total & 11{,}058 \\
\quad Indoor / Outdoor & 5{,}806 / 5{,}252 \\
\quad Drone / Manual & 7{,}051 / 4{,}007 \\
\midrule
\multicolumn{2}{l}{\textbf{Splits}} \\
\quad Train/Val (drone + manual) & 9{,}953 \\
\quad Test (drone only) & 1{,}105 \\
\midrule
\multicolumn{2}{l}{\textbf{Generated planar samples}} \\
\quad From manual ($10\times$) & 40{,}070 \\
\quad From drone ($20\times$) & 141{,}020 \\
\quad \textbf{Total} & \textbf{181{,}090} \\
\bottomrule
\end{tabular}
\end{table}

\subsection{Image Normalization and Augmentation}

Images are normalized to the range $[-1,1]$.
We apply mild affine augmentations (translation $\pm 10\%$, rotation $\pm 15^\circ$, scaling $\pm 10\%$) to increase robustness.
These augmentations are applied after projection, so that they mimic realistic camera perturbations without changing the yaw label.

\section{Learning Approach}

This section formalizes the yaw prediction task, describes the CNN architectures evaluated, and specifies the training procedure.

\subsection{Problem Formulation}

We predict a continuous yaw command from a single planar front view.
Let $x$ denote a planar RGB image and $f_\theta(\cdot)$ a CNN regressor with parameters $\theta$.
The network outputs $\hat{y} = f_\theta(x) \in [-1,1]$, which is linearly mapped back to the operational yaw range $[30^\circ,150^\circ]$.
Given a training set $\{(x_i, y_i)\}_{i=1}^{N}$ of planar images and normalized yaw labels $y_i \in [-1,1]$, we obtain $\theta$ by minimizing the mean squared error between predicted and ground-truth yaw:
\begin{equation}
\theta^* = \arg\min_{\theta} \frac{1}{N} \sum_{i=1}^{N} \left( f_\theta(x_i) - y_i \right)^2.
\end{equation}

\subsection{Architectures}

We start from a compact CNN architecture used as a baseline in prior trail-following~\cite{cnnpaper} work and adapt it from classification (three classes) to regression (single scalar output).
We evaluate nine variants grouped into five design families:
(i)~tanh baseline regressor,
(ii)~ReLU activations,
(iii)~ReLU with dropout (several dropout rates),
(iv)~ReLU with batch normalization,
(v)~leaky-ReLU (several negative slopes).

\subsection{Training Procedure}

We use supervised learning with the MSE loss defined in Section~V-A.
Optimization is done with stochastic gradient descent (SGD) with initial learning rate $5\times 10^{-3}$ and an exponential decay factor of $0.95$ per epoch.
Training runs for 100 epochs with a small batch size of 3; small-batch regimes can improve generalization and provide an implicit regularization effect~\cite{masters2018revisitingsmallbatchtraining}.
The dynamically generated dataset is split into 90\% training and 10\% validation.

\section{Experiments and Results}

We evaluate the proposed pipeline in three stages: a classification sanity check, a systematic regression comparison across architecture variants, and a semi-autonomous real-world flight test.

\subsection{Generator Validation via Classification}

As a sanity check for the dataset generator, we first reproduced the three-class (left/straight/right) classification setup of Giusti et al.~\cite{cnnpaper}.
On the held-out test set, the best run achieved 0.9201 accuracy (mean over 10 runs: 0.9099).
This indicates that the on-the-fly generation produces consistent training data and that the dataset supports learning of direction-related visual cues.

\subsection{Regression Performance}

For yaw regression, each model variant was trained 10 times with different random seeds.
The best-performing variant on the held-out test set was a leaky-ReLU regressor (negative slope $\alpha=0.05$), achieving an $R^2$ of 0.8074.
In the network's internal range $[-1,1]$, the best model yields MSE 0.0649 and RMSE 0.2548.
After mapping to degrees $[30^\circ,150^\circ]$, the same model achieves:
$15.28^\circ$ RMSE and $10.146^\circ$ MAE.

\begin{figure}[htbp]
\centerline{\includegraphics[width=0.9\columnwidth]{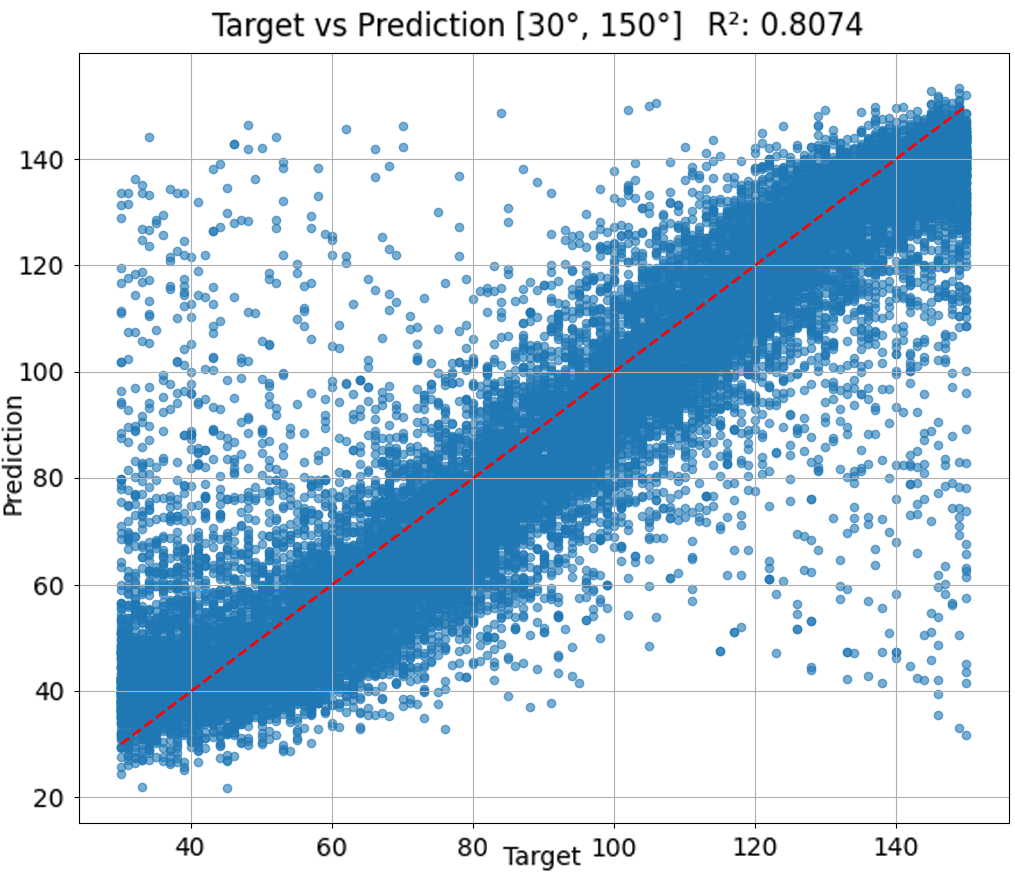}}
\caption{Predicted vs.\ target yaw values (degrees) with ideal diagonal reference.}
\label{fig:r2}
\end{figure}

The predicted-versus-target relationship in degrees is shown in
Fig.~\ref{fig:r2}, illustrating strong linear agreement across
the operational range. The predictions stay centered on the ideal diagonal with
no strong systematic bias, but the prediction error grows toward both ends of
the range: when the drone faces a wall, the view does not reveal whether the
wall lies to its left or right, making these yaw angles ambiguous and producing
the heavier tails seen at the extremes. To reduce the resulting boundary
outliers, we restricted the data sampler's field of view to the unambiguous
$[30^\circ, 150^\circ]$ region (green in Fig.~\ref{fig:winkelderdrohne}) and excluded the
red zones near the walls from training.

\begin{table}[htbp]
\caption{Best Test-Set Results Across Architecture Variants (Yaw Regression)}
\begin{center}
\begin{tabular}{|l|c|c|c|c|}
\hline
\textbf{Architecture} & \textbf{MSE} & \textbf{RMSE} & \textbf{MAE} & $\mathbf{R^2}$ \\
\hline
Tanh regressor (baseline) & 0.0772 & 0.2778 & 0.1877 & 0.7710 \\
\hline
ReLU regressor & 0.0670 & 0.2588 & 0.1761 & 0.8013 \\
\hline
ReLU + Dropout ($p\!=\!0.2$) & 0.0843 & 0.2904 & 0.2082 & 0.7498 \\
\hline
ReLU + Dropout ($p\!=\!0.3$) & 0.1159 & 0.3405 & 0.2555 & 0.6560 \\
\hline
ReLU + Dropout ($p\!=\!0.5$) & 0.1483 & 0.3851 & 0.2922 & 0.5600 \\
\hline
ReLU + Batch Normalization & 0.0779 & 0.2790 & 0.1838 & 0.7690 \\
\hline
Leaky-ReLU ($\alpha\!=\!0.01$) & 0.0655 & 0.2559 & 0.1716 & 0.8057 \\
\hline

\rowcolor{gray!20}
\textbf{Leaky-ReLU ($\boldsymbol{\alpha\!=\!0.05}$)} & \textbf{0.0649} & \textbf{0.2548} & \textbf{0.1691} & \textbf{0.8074} \\
\hline
Leaky-ReLU ($\alpha\!=\!0.1$) & 0.0650 & 0.2549 & 0.1692 & 0.8072 \\
\hline
\end{tabular}
\label{tab:regression-results}
\end{center}
\end{table}

\subsection{Ablation and Architecture Effects}

Replacing tanh activations with ReLU consistently improved validation loss and test metrics~\cite{pearsonOptimized}.
In contrast, adding dropout in the tested configurations degraded performance, suggesting that the baseline architecture and dataset already provide substantial regularization and that additional dropout can lead to underfitting.
Leaky-ReLU achieved the best overall test performance among the compared variants.

\subsection{Error Analysis}

\begin{figure*}[htbp]
\centering
\includegraphics[width=\textwidth]{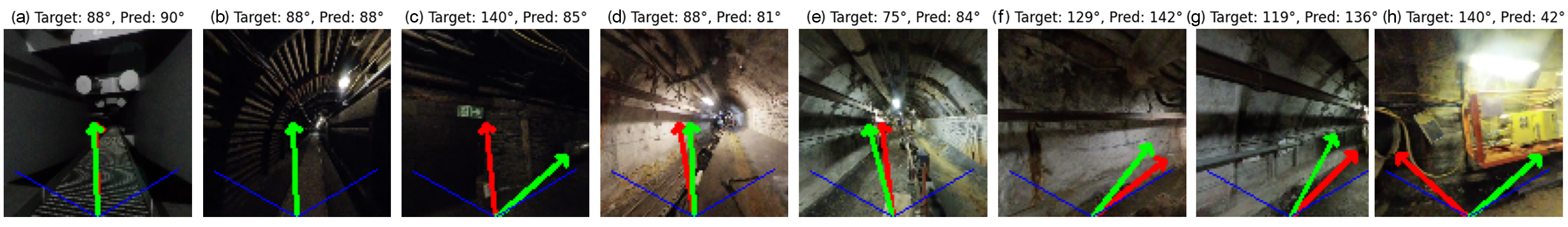}
\caption{Qualitative examples on the held-out test set: ground-truth
yaw (green) vs.\ predicted yaw (red) overlaid on planar projections.
More are available on GitHub\textsuperscript{2}.}
\label{fig:qualitative}
\end{figure*}

We observed that prediction errors are not uniformly distributed across scenes.
Fig.~\ref{fig:qualitative} summarizes representative predictions across some test
scenarios.
In most cases the predicted angle closely matches the target, resulting in
correct, collision-free navigation, as in~(a) and~(b).
In~(d)--(g), the small discrepancies arise from inaccuracies in the target angles
introduced by manual flight during dataset creation rather than from prediction
error; in several of these cases the network prediction is in fact closer to the
ideal flight direction than the labeled target, indicating that the model can
compensate for mislabeled data.

Overall, the largest prediction errors stem from two distinct failure modes.
The first is photometric: in~(h), high-dynamic-range lighting and strong specular
reflections lead the model to misinterpret bright regions as traversable openings,
causing it to predict an alternative (though potentially still valid) escape
direction.
The second is geometric and is not caused by reflections or glare: in~(c), the
scene is texture-poor and contains multiple visually plausible escape directions
of comparable openness, with no specular artifact present in the input view.
Here the network commits to a different but still geometrically valid opening,
reflecting the inherent multimodality of the underlying yaw distribution in such
scenes.
A single-frame regressor trained with an MSE objective cannot represent this
multimodality and tends to collapse onto a single mode, which can produce large
angular errors against a single ground-truth label even when the predicted
direction would still be collision-free.

\subsection{Inference Benchmark}

To evaluate suitability for robotics-relevant compute platforms, we measured inference throughput.
A total of 1001 forward passes were executed, where the first pass was used for warm-up.
The reported runtime corresponds to the average over the remaining 1000 runs.
All benchmarks were conducted using both PyTorch and ONNX runtimes.

\begin{table}[htbp]
\caption{Inference Benchmark Overview (Model Forward Pass)}
\begin{center}
\begin{tabular}{|l|c|c|}
\hline
\textbf{Hardware} & \textbf{Avg.\ Time (ms)} & \textbf{Throughput (Hz)} \\
\hline
High-end workstation GPU & 0.166--0.498 & 2008--6024 \\
\hline
Embedded GPU (Jetson Orin) & 1.486--1.528 & 654--673 \\
\hline
Low-power CPU (RPi 4 class) & 45--112 & 9--22 \\
\hline
\end{tabular}
\label{tab:benchmark}
\end{center}
\end{table}

\subsection{Real-World Semi-Autonomous Test}

A real-world semi-autonomous test was conducted using a DJI Tello EDU micro-drone.
We emphasize that this test is intended as a \emph{policy validation} rather than as a deployment demonstration: its purpose is to confirm that a network trained purely on dynamically generated planar crops of equirectangular footage produces actionable yaw commands on a previously unseen front-facing camera stream from a different platform.
The custom recording drone introduced in Section~III-A could not be used for this closed-loop test because it is deliberately designed as an \emph{invisible} recording platform: its entire payload budget is allocated to the 360$^\circ$ camera, leaving no capacity for an onboard flight computer such as a Jetson Orin or comparable embedded GPU.
The platform's ArduPilot flight controller exposes a MAVLink interface but has no onboard companion compute, no indoor position hold (GPS-only state estimation), and a proprietary analog FPV camera (Caddx Vista) that cannot be tapped for real-time inference. The DJI Tello EDU, by contrast, offers a documented Python SDK over Wi-Fi, an accessible front-camera stream, and stable indoor hover, which made it the most direct way to close the loop and validate the learned policy under laboratory conditions.
The benchmark platforms in Table~\ref{tab:benchmark} were selected because they represent the embedded compute classes typically integrated into micro-UAVs of the kind targeted by this work, and the measured throughput confirms that onboard inference is feasible on the hardware that future iterations of our platform will carry.

\begin{figure}[htbp]
\centerline{\includegraphics[width=0.99\columnwidth]{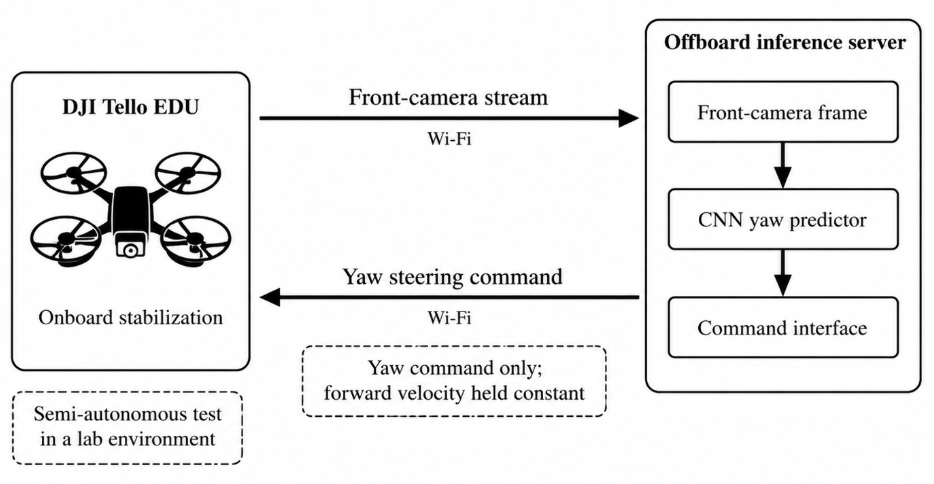}}
\caption{Offboard semi-autonomous yaw-control loop used in the real-world validation. The DJI Tello EDU streams front-camera images to an external server over Wi-Fi. The server performs CNN-based yaw prediction and sends yaw steering commands back to the drone, while forward velocity is kept constant.}
\label{fig:tello-loop}
\end{figure}

The front-camera stream of the Tello was sent via Wi-Fi to an external computer running the trained network, and the predicted yaw command was sent back as a control input while forward motion was kept constant (Fig.~\ref{fig:tello-loop}).
A previously unseen indoor lab environment was successfully traversed, including passing through a doorway without collision.

\section{Conclusion and Future Work}
The primary contribution of this paper is the acquisition of a novel real-world dataset, coupled with the implementation of an initial learning pipeline. This pipeline is designed to predict yaw commands from planar images derived from 360$^\circ$ recordings. Furthermore, we emphasize that data from open environments is significantly more valuable---and challenging to obtain---than data from restricted settings, such as industrial facilities or racing tracks.
The best model reached 15.28$^\circ$ RMSE and 10.146$^\circ$ MAE on held-out test data and demonstrated feasibility in a semi-autonomous real-world test. Our results suggest that camera-only yaw control is feasible on micro-drones when trained on representative, scenario diverse data. Replacing tanh with ReLU improved regression performance, while adding dropout degraded results, indicating possible over-regularization. Leaky-ReLU yielded the strongest overall test performance. The dominant practical limitation is robustness to illumination extremes and specular reflections.
A further structural limitation is that the single-frame reactive policy has no temporal memory: without a history of past observations, the controller cannot distinguish previously visited dead ends from unexplored corridors and may revisit regions that only appear open from a single viewpoint. Given the safety-critical nature of autonomous flight, future systems should combine learned yaw prediction with additional safeguards such as uncertainty estimation, conservative fallback behaviors, or complementary sensing.
Needless to say a lot of work remains to be done. We already build a more advanced small drone with four 180$^\circ$ fisheye cameras which will be presented in an upcoming paper. Predicting more than the yaw angle is mandatory. Future work includes: (i) increasing dataset diversity and size, (ii) targeted augmentation for brightness/contrast and
glare, (iii) using additional sensor signals (e.g., IMU) to correct potentially noisy labels, (iv) transfer learning using modern pretrained vision transformer backbones, (v) investigating
domain adaptation strategies, including fine-tuning on small post-disaster datasets and synthetic augmentation of visibility-degrading conditions (smoke, dust, water spray), to assess generalization beyond the training distribution.
Both, the dataset\footnote{\url{https://huggingface.co/datasets/RoblabWhGe/adige-360}} and the source code\footnote{\url{https://github.com/RoblabWh/adige-360}} for the training are public available.

\section*{Acknowledgment}
This work is funded by the Federal Ministry of Research, Technology and Space (BMFTR) under grant, 13N16478 (E-DRZ), cf. https://rettungsrobotik.de.

We further acknowledge the following institutions for granting access to their facilities for data collection: Deutsches Bergbau-Museum Bochum, Feuerwehr Dortmund 37 5 (Ausbildungszentrum), LWL-Museum Henrichshütte Hattingen, LWL-Museum Zeche Nachtigall Witten, LWL-Museum Zeche Zollern Dortmund, MS Kartcenter Hattingen, Straßen.NRW (Niederrheinbrücke Wesel), and Trainingsbergwerk Recklinghausen.

\bibliographystyle{IEEEtran}
\bibliography{aim2026}

\end{document}